\documentclass[11pt,oneside,dvipsnames,article]{memoir}
\usepackage[utf8]{inputenc}

\counterwithout{section}{chapter}
\setsecnumdepth{subsection}

\settypeblocksize{8in}{5.5in}{*}
\setulmargins{*}{*}{1.1}
\setlrmargins{*}{*}{1}
\setheadfoot{\onelineskip}{2\onelineskip}
\setheaderspaces{*}{2\onelineskip}{*}
\checkandfixthelayout

\usepackage{amsmath}
\usepackage{amssymb}
\usepackage{commath}

\usepackage{booktabs}
\usepackage{multirow}
\usepackage{pgfplots}

\usepackage[round]{natbib}

\usepackage[colorlinks=true,linkcolor=Brown,citecolor=Brown,filecolor=Brown,urlcolor=Brown]{hyperref}
\usepackage[all]{hypcap}

\pgfplotsset{compat=1.14}

\newcommand{\eps}{\varepsilon}
\newcommand{\lam}{\lambda}
\newcommand{\reals}{\mathbb{R}}
\newcommand{\normal}{\mathcal{N}}

\newcommand{\half}{\tfrac{1}{2}}
\newcommand{\adj}[1]{\overline{#1}}

\newcommand{\tr}[1]{{#1}^\mathsf{T}}

\DeclareMathOperator{\KL}{KL}
\DeclareMathOperator{\JS}{JS}

\newcommand{\divergence}[2]{(#1\,\|\,#2)}
\newcommand{\KLD}[2]{\KL\divergence{#1}{#2}}

\newcommand{\sref}[1]{\S\ref{#1}}
\newcommand{\tabref}[1]{Table~\ref{#1}}
\newcommand{\Eqref}[1]{(\ref{#1})}

\title{Properties of f-divergences and f-GAN training}
\author{%
  Matt Shannon \\
  Google Research \\
  \texttt{mattshannon@google.com} \\
}
\date{August 2020}

\begin{document}

\maketitle
\begin{abstract}
In this technical report we describe some properties of f-divergences and f-GAN training.
We present an elementary derivation of the f-divergence lower bounds which form the basis of f-GAN
training.
We derive informative but perhaps underappreciated properties of f-divergences and f-GAN training,
including a gradient matching property and the fact that all f-divergences agree up to an overall
scale factor on the divergence between nearby distributions.
We provide detailed expressions for computing various common f-divergences and their variational
lower bounds.
Finally, based on our reformulation, we slightly generalize f-GAN training in a way that may
improve its stability.
\end{abstract}

\section{The family of f-divergences}
\label{sec:f-def}
We start by reviewing the definition of an \emph{f-divergence} (or \emph{$\phi$-divergence})
\citep{csiszar1967information,ali1966general} and
establishing some basic properties.

\subsection{Definition}
Given a strictly convex twice continuously differentiable function $f : \reals_{> 0} \to \reals$,
the \emph{$f$-divergence} between probability distributions with densities%
\footnote{%
  Most results also hold for ``discrete'' probability distributions.
}
$p$ and $q$ over $\reals^K$ is defined as
\begin{equation}
  \label{eq:f-def}
  D_f(p, q) = \int q(x) f\left(\frac{p(x)}{q(x)}\right) \dif x
\end{equation}
For simplicity, we assume the probability distributions $p$ and $q$ are suitably nice,
e.g.\ absolutely continuous with respect to the Lebesgue measure on $\reals^K$,
$p(x), q(x) > 0$ for $x \in \reals^K$, and $p$ and $q$ continuously differentiable.
These assumptions are discussed in \sref{sec:assumptions}.
We refer to $f$ as the \emph{defining function} of the divergence $D_f$.

There is some redundancy in the above definition.
Adding an affine-linear term to the defining function just adds a constant to the divergence:
if $g(u) = f(u) + a + b u$ for $a, b \in \reals$ then $D_g(p, q) = D_f(p, q) + a + b$
for all $p$ and $q$.
Typically we do not care about an overall additive shift and would regard $D_f$ and $D_g$
as essentially the same divergence.
Thus we have identified two unnecessary degrees of freedom in the specification of the
defining function, and these may be removed by fixing $f(1) = f'(1) = 0$.
This results in no loss of generality since any defining function can be put in this form by
adding a suitable affine-linear term.%
\footnote{%
  This affine-linear term also does not affect the various bounds and finite sample
  approximations derived below, as long as the reparameterization trick is used for the
  generator gradient as is standard practice.
  In the discrete case or if other finite sample approximation such as naive REINFORCE
  is used then adding an affine-linear term to $f$ may affect the variance of the finite
  sample approximation of the generator gradient.
}
From here on we assume $f(1) = f'(1) = 0$ as part of our definition of an f-divergence.
The choice $f(1) = 0$ ensures $D_f(p, q) = 0$ when $p = q$.
The choice $f'(1) = 0$ has the ancillary benefit of \Eqref{eq:f-def} being non-negative
even if $p$ and $q$ are positive functions that do not integrate to one.

There is also a multiplicative degree of freedom in the defining function that is often
irrelevant:
multiplying the defining function by a constant $k > 0$ just multiplies the divergence by the
same constant.
Removing this superficial source of variation makes different f-divergences easier to compare.
We refer to an f-divergence as being \emph{canonical} if $f''(1) = 1$.
Any f-divergence can be made canonical by scaling appropriately.
Intuitively, fixing $f''(1)$ corresponds to fixing the behavior of the f-divergence in the
region $u \approx 1$, corresponding to $p \approx q$.
This will be made precise below.

\subsection{Properties}
f-divergences satisfy several mathematical properties:
\begin{itemize}
\item
  $D_f$ is linear in $f$.
\item
  $D_f(p, q) \geq 0$ for all distributions $p$ and $q$ with equality iff $p = q$.
  This justifies referring to $D_f$ as a divergence.
\item
  $D_f$ uniquely determines $f$.
\item
  All f-divergences agree up to an overall scale factor on the divergence between nearby
  distributions: If $p \approx q$ then $D_f(p, q) \approx f''(1) \KLD{p}{q}$ to second
  order.
  In particular all canonical f-divergences agree when $p \approx q$.
\item
  For many common f-divergences, $f''$ has a simpler algebraic form than $f$ and is
  easier to work with.
\end{itemize}

Linearity is straightforward to verify.
If $D_f$ and $D_g$ are two f-divergences and $k > 0$ then $D_{f + g} = D_f + D_g$
and $D_{(k f)} = k D_f$.
If $f$ and $g$ are strictly convex then so is $f + g$ and $k f$, and
$(f + g)(1) = f(1) + g(1) = 0$ and $(k f)(1) = k f(1) = 0$, and similarly for $f'$,
so $D_{f + g}$ and $D_{k f}$ are valid f-divergences.

The non-negativity of $D_f$ follows from the convexity of $f$.
Since $f(1) = f'(1) = 0$ and $f$ is strictly convex, $f(u) \geq 0$ with equality iff
$u = 1$.
Thus
\begin{equation}
  \label{eq:f-non-neg}
  \int q(x) f\left(\frac{p(x)}{q(x)}\right) \dif x \geq 0
\end{equation}
Thus $D_f(p, q) \geq 0$ for all $p$ and $q$.
In general, if $\int g(x) \dif x = 0$ for a continuous integrable non-negative function
$g : \reals^K \to \reals$ then $g(x) = 0$ for all $x$.
Applying this to $g(x) = q(x) f(p(x) / q(x))$ we see that we have equality in
\Eqref{eq:f-non-neg} iff $p(x) / q(x) = 1$ for all $x$.
Thus $D_f(p, q) = 0$ implies $p = q$.
The non-negativity of $D_f$ can also be seen by plugging the constant function
$u(x) = 1$ into \Eqref{eq:f-bound}.

We now show that $D_f$ completely determines $f$.
Suppose $D_f = D_g$.
We wish to show that $f = g$.
Consider first the discrete case where $p$ and $q$ are distributions over a two-point set
$\{ 0, 1 \}$.
Given $u > 1$, choose $p_u$ and $q_u$ such that $p_u(0) / q_u(0) = u$ and
$p_u(1) / q_u(1) = \half$.
It is straightforward to show that such $p_u$ and $q_u$ exist and are given by
\begin{align}
  p_u(0) &= \frac{u}{2 u - 1}
  &
  q_u(0) &= \frac{1}{2 u - 1}
  \\
  p_u(1) &= \frac{u - 1}{2 u - 1}
  &
  q_u(1) &= \frac{2 (u - 1)}{2 u - 1}
\end{align}
Thus
\begin{equation}
  (2 u - 1) D_f(p_u, q_u) = f(u) + 2 (u - 1) f(\half)
\end{equation}
Subtracting the equivalent equation for $g$ and using $D_f = D_g$, we see that $f$ and $g$
differ only in an affine-linear term: $g(u) = f(u) + a (u - 1)$ for all $u > 1$ for some
$a \in \reals$.
Therefore $g'(u) = f'(u) + a$ for $u > 1$, and taking the limit as $u \to 1$ we have
$g'(1) = f'(1) + a$.
But $f'(1) = g'(1) = 0$, so $a = 0$, so $f(u) = g(u)$ for $u > 1$.
A similar argument applies for $0 < u < 1$.
In this case we choose $p_u(0) / q_u(0) = u$ and $p_u(1) / q_u(1) = 2$, and we have
\begin{align}
  p_u(0) &= \frac{u}{2 - u}
  &
  q_u(0) &= \frac{1}{2 - u}
  \\
  p_u(1) &= \frac{2 (1 - u)}{2 - u}
  &
  q_u(1) &= \frac{1 - u}{2 - u}
\end{align}
and
\begin{equation}
  (2 - u) D_f(p_u, q_u) = f(u) + (1 - u) f(2)
\end{equation}
From here we apply the same argument as above.
Thus $f(u) = g(u)$ for $0 < u < 1$.
Combining this with the above result for $u > 1$ and the fact that $f(1) = g(1)$,
we have $f = g$ as desired.
For the continuous case where $p$ and $q$ are densities over $\reals^K$, we reduce this
to the discrete case by considering mixtures of Gaussians with shrinking covariances.
Consider the $u > 1$ case.
Fix any two distinct points $\mu_0, \mu_1 \in \reals^K$.
Let
$\tilde{p}_{u \sigma}(x)
    = p_u(0)\,\normal(x; \mu_0, \sigma^2 I) + p_u(1)\,\normal(x; \mu_1, \sigma^2 I)$
and
$\tilde{q}_{u \sigma}(x)
    = q_u(0)\,\normal(x; \mu_0, \sigma^2 I) + q_u(1)\,\normal(x; \mu_1, \sigma^2 I)$
where $p_u$ and $q_u$ are as specified for the discrete $u > 1$ case above.
As $\sigma \to 0$, $D_f(\tilde{p}_{u \sigma}, \tilde{q}_{u \sigma}) \to D_f(p_u, q_u)$, so
$(2 u - 1) D_f(\tilde{p}_{u \sigma}, \tilde{q}_{u \sigma}) \to f(u) + 2(u - 1) f(\half)$,
and similarly for $g$.
But $D_f = D_g$, so we have the same quantity tending to two limits, so the limits must be
equal.
Thus $g(u) = f(u) + a (u - 1)$ for some $a \in \reals$ and the remainder of the argument is
as above.
A similar argument applies for $0 < u < 1$.
Thus $f = g$ also holds in the continuous case.

Different f-divergences may behave very differently when $p$ and $q$ are far apart but
are essentially identical when $q \approx p$.
One way to make this precise is to consider a parametric family $\{ q_\lam: \lam \in \Lambda \}$
of densities.
A Taylor expansion of $D_f\left(q_\lam, q_{\lam + \eps v}\right)$ in terms of $\epsilon$ shows
that
\begin{equation}
  \label{eq:f-div-p-approx-q}
  D_f\left(q_\lam, q_{\lam + \eps v}\right) =
    \half \eps^2 f''(1) \tr{v} F(\lam) v + O(\eps^3)
\end{equation}
where $\eps \in \reals$, $v \in \reals^K$, and
\begin{equation}
  F_{i j}(\lam) = \int q_\lam(x)
    \left(\tpd{}{\lam_i}\log q_\lam(x)\right) \left(\tpd{}{\lam_j}\log q_\lam(x)\right)
    \dif x
\end{equation}
is the Fisher information matrix of the parametric family, also known as the Fisher metric.
Thus all f-divergences agree up to a constant factor on the divergence between two nearby
distributions, and they are all just scaled versions of the Fisher distance in this regime.
Alternatively this may be stated in the non-parametric form
\begin{equation}
  \label{eq:f-div-p-approx-q-non-parametric}
  D_f\left(q, q + \eps v\right) =
    \half \eps^2 f''(1) \int \frac{\left(v(x)\right)^2}{q(x)} \dif x + O(\eps^3)
\end{equation}
where $v : \reals^K \to \reals$ satisfies $\int v(x) \dif x = 0$.
Informally we may state this as
\begin{equation}
  \label{eq:f-div-p-approx-q-simple}
  D_f(p, q) \approx \half f''(1) \int \frac{\left(p(x) - q(x)\right)^2}{p(x)} \dif x
\end{equation}
Thus all f-divergences agree up to a constant factor on the divergence between nearby
distributions.

The fact that $f''$ has a simpler algebraic form than $f$ for many common f-divergences
will be seen when we consider specific f-divergences below.
Considering $f''$ is natural for a number of reasons.
Adding an affine-linear term to $f$ does not change $f''$.
Thus even without the constraints $f(1) = f'(1) = 0$, $f''$ would not have the two
unnecessary degrees of freedom mentioned above and $D_f$ would uniquely determine $f''$.
Strict convexity of $f$ corresponds to the simple condition $f''(u) > 0$ for all $u$.
We will also see below that various gradients of $D_f$ and of its lower bound $E_f$ depend
on $f$ only through $f''$.
Considering $f''$ also provides a simple view of how various f-divergences are related.
For example, three common symmetric divergences are the Jensen-Shannon, squared Hellinger
and Jeffreys divergences.
In the $f''$ domain, these may all be viewed as forms of average of KL and reverse KL,
specifically as the harmonic mean, geometric mean and arithmetic mean respectively.

\section{Variational divergence estimation}
\label{sec:vde}
f-GANs are based on an elegant way to estimate the f-divergence between two distributions
given only samples from the two distributions \citep{nguyen2010estimating}.
In this section we review this approach to \emph{variational divergence estimation}.
We provide a simple and easy to understand derivation involving elementary facts about
convex functions.
At the end of this section we discuss how our derivation and notation
relates to that of the original f-GAN paper \citep{nowozin_f-gan:_2016}.

\subsection{Variational lower bound}
\begin{figure}
  \centering
  \begin{tikzpicture}
    \begin{axis}[axis lines=middle,xlabel=$u$,ymin=-0.5]
      \addplot[domain=0:2.4,samples=100]{(x - 1)^2};
      \addlegendentry{$f(u)$}

      \addplot[domain=0:2.4,dashed]{x - 1.25};
      \addlegendentry{Tangent line}
    \end{axis}
  \end{tikzpicture}
  \caption{%
    A convex function $f : \reals_{> 0} \to \reals$ and a tangent line.
    The variational bound used by f-GANs is based on the fact that a convex function $f$
    lies at or above its tangent lines.
  }
  \label{fig:tangent-lines}
\end{figure}
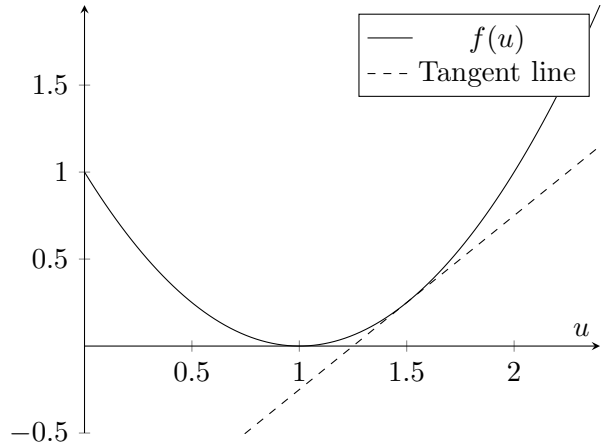
We first derive a variational lower bound on the f-divergence $D_f$.
Since $f$ is strictly convex, its graph lies at or above any of its tangent lines and
only touches in one place.
That is, for $k, u > 0$,
\begin{equation}
  f(k) \geq f(u) + (k - u) f'(u) = k f'(u) - \left[u f'(u) - f(u)\right]
\end{equation}
with equality iff $k = u$.
This inequality is illustrated in \figref{fig:tangent-lines}.
Substituting $p(x) / q(x)$ for $k$ and $u(x)$ for $u$, for any continuously differentiable
function $u : \reals^K \to \reals_{> 0}$ we obtain
\begin{equation}
  \label{eq:f-bound}
  D_f(p, q) \geq
    \int p(x) f'(u(x)) \dif x - \int q(x) \left[u(x) f'(u(x)) - f(u(x))\right] \dif x
\end{equation}
with equality iff $u = u^*$, where $u^*(x) = p(x) / q(x)$.
The function $u$ is referred to as the \emph{critic}.
It will be helpful to have a concise notation for this bound.
Writing $u(x) = \exp(d(x))$ without loss of generality, for any continuously differentiable
function $d : \reals^K \to \reals$, we have
\begin{equation}
  \label{eq:div-bound}
  D_f(p, q) \geq E_f(p, q, d)
\end{equation}
with equality iff $d = d^*$, where
\begin{align}
  \label{eq:f-bound-def}
  E_f(p, q, d) &= \int p(x) a_f(d(x)) \dif x - \int q(x) b_f(d(x)) \dif x
  \\
  a_f(d) &= f'(\exp(d))
  \\
  b_f(d) &= \exp(d) f'(\exp(d)) - f(\exp(d))
  \\
  d^*(x) &= \log p(x) - \log q(x)
\end{align}
Note that both $a_f$ and $b_f$ are linear in $f$.
Their derivatives $a_f'(\log u) = u f''(u)$ and $b_f'(\log u) = u^2 f''(u)$
depend on $f$ only through $f''$.
Note that $b_f'(d) = a_f'(d) \exp(d)$.

\subsection{Formulation of variational divergence estimation}
The bound \Eqref{eq:div-bound} leads naturally to \emph{variational divergence estimation}.
The $f$-divergence between $p$ and $q$ can be estimated by maximizing $E_f$ with respect
to $d$ \citep{nguyen2010estimating}.
Conveniently $E_f$ is expressed in terms of expectations and may be approximately
computed and maximized with respect to $d$ using only samples from $p$ and $q$.
Ultimately this property derives from the fact that the tangent lines of $f(u)$ are
affine-linear in $u$, and the constant term leads to expectations with respect to $q$
and the linear term leads to expectations with respect to $p$.
If we parameterize $d$ as a neural net $d_\nu$ with parameters $\nu$ then we can
approximate the divergence by maximizing $E_f(p, q, d_\nu)$ with respect to $\nu$.
This does not compute the exact divergence for several reasons:
there is no guarantee that $d^*$ lies in the family $\{ d_\nu : \nu \}$ of functions
representable by the neural net;
gradient-based optimization may find a local but not global minimum;
and we have to be careful not to overfit given a finite set of samples from $p$
(typically $q$ is a model and so we have access to arbitrarily many samples).
However we hope for sufficiently flexible neural nets and careful optimization that
the approximation will be close.

\subsection{Expressions for common f-divergences}
\label{sec:divergences-more}
In this section we express several common divergences in terms of \Eqref{eq:f-def} and
the variational lower bound \Eqref{eq:f-bound-def}.
For each f-divergence, we give explicit expressions for
$f$, $f''$, $D_f$, $E_f$, $a_f$, $b_f$, $a_f'$ and $b_f'$.
We also list the \emph{tail weights}, which are related to the asymptotic behavior of
$f''(u)$ as $u \to 0$ and $u \to \infty$ and which determine the key qualitative
properties of an f-divergence \citep{shannon2020divergences}.
We mention below how some of the divergences are related to each other by \emph{softening}
\citep{shannon2020divergences}.
The \emph{p-softening} of a divergence $D$ is the divergence $\tilde{D}$ given by
$\tilde{D}(p, q) = 4 D(\half p + \half q, q)$.
Similarly the \emph{q-softening} is given by $\tilde{D}(p, q) = 4 D(p, \half p + \half q)$.
The factor of $4$ here ensures that $\tilde{D}$ remains canonical
\citep{shannon2020divergences}.

The \emph{Kullback-Leibler (KL) divergence} (or \emph{I divergence} or \emph{relative entropy})
satisfies:
\begin{align}
  D_f(p, q) &= \KLD{p}{q} = \int p(x) \log\frac{p(x)}{q(x)} \dif x
  \\
  E_f(p, q, d) &= 1 + \int p(x) d(x) \dif x - \int q(x) \exp(d(x)) \dif x
  \\
  f(u) &= u \log u - u + 1
  \\
  f''(u) &= u^{-1}
  \\
  a_f(d) &= d
  \\
  b_f(d) &= \exp(d) - 1
  \\
  a_f'(d) &= 1
  \\
  b_f'(d) &= \exp(d)
\end{align}
The KL divergence has $(1, 2)$ tail weights.
The KL divergence defining function is sometimes given as $f(u) = u \log u$.
The additional affine-linear term in our expression for $f$ is due to the constraints
$f(1) = f'(1) = 0$.
We note in passing that this precisely corresponds to what is sometimes referred to as the
\emph{generalized KL divergence}
$D(p, q) = \int p(x) \log\frac{p(x)}{q(x)} \dif x - \int p(x) \dif x + \int q(x) \dif x$.
This has the property that $D(p, q) \geq 0$ with equality iff $p = q$ even if we remove
the constraint that $p$ and $q$ be valid densities that integrate to one.

The \emph{reverse KL divergence} satisfies:
\begin{align}
  D_f(p, q) &= \KLD{q}{p} = \int q(x) \log\frac{q(x)}{p(x)} \dif x
  \\
  E_f(p, q, d) &= 1 - \int p(x) \exp(-d(x)) \dif x - \int q(x) d(x) \dif x
  \\
  f(u) &= -\log u + u - 1
  \\
  f''(u) &= u^{-2}
  \\
  a_f(d) &= 1 - \exp(-d)
  \\
  b_f(d) &= d
  \\
  a_f'(d) &= \exp(-d)
  \\
  b_f'(d) &= 1
\end{align}
The reverse KL divergence has $(2, 1)$ tail weights.
Note the explicit symmetry between the representations of KL and reverse KL in terms of
$a_f$ and $b_f$.
Their symmetric relationship is less apparent from $f$ and $f''$.
In general if $D_g(p, q) = D_f(q, p)$ then $a_g(d) = b_f(-d)$ and $b_g(d) = a_f(-d)$.

The canonicalized \emph{Jensen-Shannon divergence} \citep{lin1991divergence}
(or \emph{Jensen difference} \citep{burbea1982convexity}
or \emph{capacitory discrimination} \citep{topsoe2000some}) satisfies:
\begin{align}
  D_f(p, q) &= 4 \JS(p, q)
  \\
    &= 2 \KLD{p}{\half p + \half q} + 2 \KLD{q}{\half p + \half q}
  \\
    &= 4 \log 2 + 2 \int p(x) \log\frac{p(x)}{p(x) + q(x)} \dif x
      + 2 \int q(x) \log\frac{q(x)}{p(x) + q(x)} \dif x
  \\
  E_f(p, q, d)
    &= 4 \log 2 + 2 \int p(x) \log\sigma(d(x)) \dif x + 2 \int q(x) \log\sigma(-d(x)) \dif x
  \\
  f(u) &= 2 u \log u - 2 (u + 1) \log (u + 1) + 2 u \log 2 + 2 \log 2
  \\
  f''(u) &= \frac{2}{u (u + 1)}
  \\
  a_f(d) &= 2 \log\sigma(d) + 2 \log 2
  \\
  b_f(d) &= -2 \log\sigma(-d) - 2 \log 2
  \\
  a_f'(d) &= 2 \sigma(-d)
  \\
  b_f'(d) &= 2 \sigma(d)
\end{align}
Here $\JS$ is the conventional, non-canonical Jensen-Shannon divergence with $f''(1) = 1/4$.
The Jensen-Shannon divergence has $(1, 1)$ tail weights.
Its $f''(u)$ is the harmonic mean of $f''(u)$ for KL and $f''(u)$ for reverse KL.
The square root of the Jensen-Shannon divergence defines a metric on the space of
probability distributions \citep{endres2003new,osterreicher2003new}.

The canonicalized \emph{squared Hellinger distance} (closely related to the
\emph{Freeman-Tukey statistic} for hypothesis testing) satisfies:
\begin{align}
  D_f(p, q) &= 2 \int \left( \sqrt{p(x)} - \sqrt{q(x)} \right)^2 \dif x
  \\
    &= 4 - 4 \int \sqrt{p(x) q(x)} \dif x
  \\
  E_f(p, q, d)
    &= 4 - 2 \int p(x) \exp\left( -\half d(x) \right) \dif x
      - 2 \int q(x) \exp\left( \half d(x) \right) \dif x
  \\
  f(u) &= 2 (1 - \sqrt{u})^2
  \\
  f''(u) &= u^{-\frac{3}{2}}
  \\
  a_f(d) &= 2 - 2 \exp\left(-\half d\right)
  \\
  b_f(d) &= 2 \exp\left(\half d\right) - 2
  \\
  a_f'(d) &= \exp\left(-\half d\right)
  \\
  b_f'(d) &= \exp\left(\half d\right)
\end{align}
Here $\int \sqrt{p(x) q(x)} \dif x$ is known as the \emph{Bhattacharyya coefficient}.
The squared Hellinger distance has $(\frac{3}{2}, \frac{3}{2})$ tail weights.
Its $f''(u)$ is the geometric mean of $f''(u)$ for KL and $f''(u)$ for reverse KL.
The Hellinger distance defines a metric on the space of probability distributions
\citep{vajda2009metric}.

The \emph{Jeffreys divergence} (or \emph{J divergence})
$(p, q) \mapsto \half \KL(p, q) + \half \KL(q, p)$
is the arithmetic mean of the KL divergence and reverse KL divergence.
Since $f$, $f''$, $D_f$, $E_f$, $a_f$, $b_f$, $a_f'$ and $b_f'$ are all linear in $f$, they
are also all just the arithmetic mean of the corresponding quantities for KL and reverse KL
and so we do not list them separately.
The Jeffreys divergence has $(2, 2)$ tail weights.

The canonicalized \emph{squared Le Cam distance} (or \emph{squared Puri-Vincze distance}
or \emph{triangular discrimination}) satisfies:
\begin{align}
  D_f(p, q) &= 2 \Delta(p, q)
  \\
    &= \chi^2(p, \half p + \half q)
  \\
    &= \chi^2(q, \half p + \half q)
  \\
    &= \int\frac{\left(p(x) - q(x)\right)^2}{p(x) + q(x)} \dif x
  \\
  E_f(p, q, d) &=
    2
    - 4 \int p(x) \left(\sigma(-d(x))\right)^2 \dif x
    - 4 \int q(x) \left(\sigma(d(x))\right)^2 \dif x
  \\
  f(u) &= \frac{(u - 1)^2}{1 + u}
  \\
  f''(u) &= \frac{8}{(1 + u)^3}
  \\
  a_f(d) &= 1 - 4 \left(\sigma(-d)\right)^2
  \\
  b_f(d) &= 4 \left(\sigma(d)\right)^2 - 1
  \\
  a_f'(d) &= 8 \sigma(d) \left(\sigma(-d)\right)^2
  \\
  b_f'(d) &= 8 \left(\sigma(d)\right)^2 \sigma(-d)
\end{align}
Here $\Delta$ is the conventional, non-canonical squared Le Cam distance with $f''(1) = \half$,
and $\chi^2$ is the Pearson $\chi^2$ divergence defined below.
The squared Le Cam distance has $(0, 0)$ tail weights.
It is symmetric with respect to $p$ and $q$.
The canonicalized squared Le Cam distance may be obtained by q-softening the canonicalized
Pearson $\chi^2$ divergence or p-softening the canonicalized Neymann divergence.
The Le Cam distance defines a metric on the space of probability distributions
\citep{vajda2009metric}.

The canonicalized \emph{Pearson $\chi^2$ divergence} (or \emph{Kagan divergence}) satisfies:
\begin{align}
  D_f(p, q) &= \half \chi^2(p, q)
  \\
    &= \half \int\frac{\left(p(x) - q(x)\right)^2}{q(x)} \dif x
  \\
  E_f(p, q, d) &=
    -\half
    + \int p(x) \exp\left(d(x)\right) \dif x
    - \half \int q(x) \exp\left(2 d(x)\right) \dif x
  \\
  f(u) &= \half (u - 1)^2
  \\
  f''(u) &= 1
  \\
  a_f(d) &= \exp(d) - 1
  \\
  b_f(d) &= \half \exp(2 d) - \half
  \\
  a_f'(d) &= \exp(d)
  \\
  b_f'(d) &= \exp(2 d)
\end{align}
Here $\chi^2$ is the conventional, non-canonical Pearson $\chi^2$ divergence with $f''(1) = 2$.
The Pearson $\chi^2$ divergence has $(0, 3)$ tail weights.
The p-softened canonicalized Pearson $\chi^2$ divergence is itself.
The q-softened canonicalized Pearson $\chi^2$ divergence is the
canonicalized squared Le Cam distance.

The canonicalized \emph{Neymann divergence} satisfies:
\begin{align}
  D_f(p, q) &= \half \chi^2(q, p)
  \\
    &= \half \int\frac{\left(p(x) - q(x)\right)^2}{p(x)} \dif x
  \\
  E_f(p, q, d) &=
    -\half - \half \int p(x) \exp\left(-2 d(x)\right) \dif x
    + \int q(x) \exp\left(-d(x)\right) \dif x
  \\
  f(u) &= \frac{(u - 1)^2}{2 u}
  \\
  f''(u) &= u^{-3}
  \\
  a_f(d) &= \half  - \half \exp(-2 d)
  \\
  b_f(d) &= 1 - \exp(-d)
  \\
  a_f'(d) &= \exp(-2 d)
  \\
  b_f'(d) &= \exp(-d)
\end{align}
The Neymann divergence has $(3, 0)$ tail weights.
It is the reverse of the Pearson $\chi^2$ divergence.
The p-softened canonicalized Neymann divergence is the
canonicalized squared Le Cam distance.
The q-softened canonicalized Neymann divergence is itself.

% TODO(mattshannon): Fill-in number for shannon2020divergences on arXiv.
The \emph{softened reverse KL divergence} \citep{shannon2020divergences} satisfies:
\begin{align}
  D_f(p, q) &= 4 \KLD{\half p + \half q}{p}
  \\
  \begin{split}
  E_f(p, q, d) &= 2 - 4 \log 2
    + 2 \int p(x) \left[ -\exp\left(-d(x)\right) - \log\sigma\left(d(x)\right) \right] \dif x
  \\
    &\quad - 2 \int q(x) \log\sigma\left(d(x)\right) \dif x
  \end{split}
  \\
  f(u) &= 2 (u + 1) \log\frac{u + 1}{u} - 4 \log 2
  \\
  f''(u) &= \frac{2}{u^2 (u + 1)}
  \\
  a_f(d) &= -2 \exp(-d) - 2 \log\sigma(d) - 2 - 2 \log 2
  \\
  b_f(d) &= 2 \log\sigma(d) + 2 \log 2
  \\
  a_f'(d) &= 2 \exp(-d) \sigma(-d)
  \\
  b_f'(d) &= 2 \sigma(-d)
\end{align}
The softened reverse KL divergence has $(2, 0)$ tail weights.
It is obtained by q-softening the reverse KL divergence.
This is the divergence approximately minimized by conventional non-saturating GAN
training \citep{shannon2020divergences}.

\begin{table}[t]
  \centering
  \begin{tabular}{l c c l}
    \toprule
    \multirow{2}{*}{divergence} & \multirow{2}{*}{$f''(u)$} & (left, right) & \\
    & & tail weights & \\
    \midrule
    KL & $u^{-1}$ & $(1, 2)$ & \\
    reverse KL & $u^{-2}$ & $(2, 1)$ & \\
    Jensen-Shannon & $\frac{2}{u (1 + u)}$ & $(1, 1)$ & (harmonic mean of KL and RKL) \\
    squared Hellinger & $u^{-\frac{3}{2}}$ & $(\frac{3}{2}, \frac{3}{2})$ &
        (geometric mean of KL and RKL) \\
    Jeffreys & $\frac{1 + u}{2 u^2}$ & $(2, 2)$ &
        (arithmetic mean of KL and RKL) \\
    squared Le Cam & $\frac{8}{(1 + u)^3}$ & $(0, 0)$ & \\
    Pearson $\chi^2$ & $1$ & $(0, 3)$ & \\
    Neymann & $u^{-3}$ & $(3, 0)$ & \\
    softened reverse KL & $\frac{2}{u^2 (1 + u)}$ & $(2, 0)$ & \\
    \bottomrule
  \end{tabular}
  \caption{%
    Concise specification of various f-divergences in terms of $f''$.
    The divergences are scaled to make them canonical ($f''(1) = 1$).
    All are low-degree rational functions of $u$ (or $\sqrt{u})$.
    Tail weights, which determine the most important qualitative properties of an
    f-divergence, are also shown \citep{shannon2020divergences}.
  }%
  \label{tab:f-dash-dash}
\end{table}
The $f''$ for various f-divergences is summarized in \tabref{tab:f-dash-dash}.
We see that $f''$ provides a particular simple and concise way to define many common
f-divergences.
These are all rational functions of $\sqrt{u}$.

\subsection{Relationship to original f-GAN formulation}
The original f-GAN paper \citep{nowozin_f-gan:_2016} phrases the results presented in
\sref{sec:vde} in terms of the \emph{Legendre transform} or \emph{Fenchel conjugate}
$f^*$ of $f$.
The two descriptions are equivalent%
\footnote{%
  Assuming $f$ is differentiable.
  This is also assumed in practice in the original f-GAN paper.
  If this was not the case then it would not be possible to train the critic using
  gradient-based optimization.
},
as can be seen by setting $T(x) = f'(u(x))$ and using the result $f^*(f'(u)) = u f'(u) - f(u)$.
We find our description helpful since it avoids having to explicitly match the domain of
$f^*$, ensures the optimal $d$ is the same for all $f$-divergences, and because the Legendre
transform is complicated for one of the divergences we consider.
An ``output activation'' was used in the original f-GAN paper to adapt the output $d$ of
the neural net to the domain of $f^*$.
This is equal to $f'(\exp(d))$, up to irrelevant additive constants, for all the divergences
we consider, and so our description also matches the original description in this respect.

\section{Variational divergence minimization}
\label{sec:vdm}
f-GANs \citep{nowozin_f-gan:_2016} generalize classic GANs to allow approximately minimizing any
f-divergence.
In this section we review and discuss the f-GAN formulation.

Consider the task of estimating a probabilistic model from data using an f-divergence.
Here $p$ is the true distribution and the goal is to minimize $l(\lam) = D_f(p, q_\lam)$
with respect to $\lam$, where $\lam \mapsto q_\lam$ is a parametric family of densities
over $\reals^K$.
We refer to $q_\lam$ as the \emph{generator}.
For implicit generative models such as typical GAN generators, the distribution $q_\lam$
is the result of a deterministic transform $g_\lam(z)$ of a stochastic latent variable
$z$.
However we do not need to assume this specific form for most of our discussion.

\subsection{Gradient matching property}
We first note that the variational divergence bound $E_f$ satisfies a convenient gradient
matching property.
This is not made explicit in the original f-GAN paper.
Denote the optimal $d$ given $p$ and $q_\lam$ by $d^*_\lam$.
We saw above that $D_f(p, q_\lam)$ and $E_f(p, q_\lam, d)$ match values at $d = d^*_\lam$.
They also match gradients with respect to the generator parameters $\lam$:
\begin{equation}
  \label{eq:grad-match}
  \dpd{}{\lam} D_f(p, q_\lam) =
    \dpd{}{\lam} E_f(p, q_\lam, d)\sVert[4]_{d = d^*_\lam}
      = -\int \left[\dpd{}{\lam} q_\lam(x)\right] b_f(d^*_\lam(x)) \dif x
\end{equation}
This follows from the fact that $E_f$ is a tight lower bound on $D_f$, similarly to
the one-dimensional result that any differentiable function $f : \reals \to \reals$
with $f(x) \geq 0$ for all $x$ and $f(0) = 0$ has $f'(0) = 0$.
We can also verify this property directly from the definitions of $D_f$ and $E_f$.

\subsection{Formulation of variational divergence minimization}
We can minimize $D_f(p, q_\lam)$ using \emph{variational divergence minimization},
maximizing $E_f(p, q_\lam, d_\nu)$ with respect to $\nu$ while minimizing it with
respect to $\lam$.
Adversarial optimization such as this lies at the heart of all flavors of GAN training.
Define $\adj{\lam}$ and $\adj{\nu}$ as
\begin{align}
  \label{eq:generator-gradient}
  \adj{\lam} &= -\dpd{}{\lam}E_f(p, q_\lam, d_\nu)
    = \int \left[\dpd{}{\lam} q_\lam(x)\right] b_f(d_\nu(x)) \dif x
  \\
  \label{eq:critic-gradient}
  \adj{\nu} &= \dpd{}{\nu}E_f(p, q_\lam, d_\nu)
\end{align}
To perform the adversarial optimization, we can feed $\adj{\lam}$ and $\adj{\nu}$ (or
in practice, stochastic approximations to them) as the gradients into any gradient-based
optimizer designed for minimization, e.g.\ stochastic gradient descent or ADAM.

The gradient matching property shows that performing very many critic updates followed
by a single generator update is a sensible learning strategy which, assuming the critic
is sufficiently flexible and amenable to optimization, essentially performs very slow
gradient-based optimization on the true divergence $D_f$ with respect to $\lam$.
However in practice performing a few critic updates for each generator update, or
simultaneous generator and critic updates, performs well, and it is easy to see that
these approaches at least have the correct fixed points in terms of Nash equilibria of $E_f$
and optima of $D_f$, subject as always to the assumption that the critic is sufficiently
richly parameterized.
Convergence properties of these schemes are investigated much more thoroughly elsewhere,
for example
\citep{nagarajan2017gradient,gulrajani2017improved,%
mescheder2017numerics,mescheder2018training,%
balduzzi2018mechanics,peng2019training},
and are not the main focus here.

\subsection{Hybrid training schemes}
There is a simple generalization of the above training procedure, which is to base the
generator gradients on $E_f$ but the critic gradients on $E_h$ for a possibly different
function $h$ \cite[Section 2.3]{poole2016improved}.
We refer to this as using \emph{hybrid $(f, h)$} gradients.
This also approximately minimizes $D_f$.
Subject as always to the assumption of a richly parameterized critic, if we perform very
many critic updates for each generator update, then the $d$ used to compute the generator
gradient will still be close to $d^*$, and so the generator gradient will be close to
the gradient of $D_f$, even though the path $d$ took to approach $d^*$ was governed by
$g$ rather than $f$.
The fixed points of the two gradients are also still correct, and so it seems reasonable
to again use more general update schemes and we might hope for similar convergence
results (not analyzed here).

Hybrid schemes may potentially be useful for stabilizing training.
For example the reverse KL generator gradient depends on $f$ only through $b_f'(d) = 1$,
so is likely to be stable with respect to minor inaccuracies in the critic, but the
reverse KL critic gradient involves $a_f'(d) = \exp(-d)$ which may lead to very large
updates if $d(x)$ is ever large and negative for real $x$.
A hybrid (reverse KL, Jensen-Shannon) scheme uses a stable update for both the generator
and critic while still approximately minimizing reverse KL.

\subsection{Low-dimensional generator support}
\label{sec:assumptions}
Many GAN generators used in practice have low-dimensional support and do not satisfy
the condition $q(x) > 0$ for all $x$ which we assumed for simplicity.
In this section we briefly discuss the implications of this for GAN training.
We argue that using generators with $q(x) > 0$ everywhere has both theoretical and
practical benefits, and so it is not unreasonable to restrict attention to this case.

The vast majority of GAN generators consist of a deterministic neural net applied to a
fixed source of noise.
Often the noise is far lower-dimensional than the output space, meaning that the set of
possible generator outputs for a given trained generator (its \emph{support} as a
probability distribution) is a low-dimensional manifold in output space.
For example, the progressive GAN generator \citep{karras2018progressive} used a
$512$-dimensional noise source and an output space with roughly $3$ million dimensions.
The natural data is often assumed to also lie on a low-dimensional manifold in output
space, but we would argue that this is almost never exactly the case in practice.
Due to sensor noise and dithering if nothing else, it is difficult to say that any
image, say, is literally \emph{impossible} in natural data.
If desired, we can even guarantee that this is the case by adding a perceptually
insignificant amount of white noise to the data.
It seems more accurate to say that the natural data lies \emph{close to} a low-dimensional
manifold, but that no output is impossible, i.e.\ $p(x) > 0$ everywhere.
The low-dimensional generator support combined with high-dimensional data distribution
support leads to several pathologies:
\begin{itemize}
\item
  The set of all possible generator outputs has probability zero under the data distribution.
\item
  With probability $1$, the generator assigns a natural image a probability density of zero.
\item
  The KL divergence between the data distribution and the generator is infinite.
\item
  The true log likelihood of natural data under the model is $-\infty$ (despite approaches
  based on Parzen windows which produce finite estimates for the log likelihood).
\item
  Essentially all f-divergences are either undefined or completely saturate with
  gradient precisely zero.
\item
  The optimal critic $d^*(x) = \log p(x) - \log q(x)$ is $\infty$ almost everywhere.
\item
  A sufficiently powerful critic can learn to distinguish generator output essentially
  perfectly by outputting ``fake'' for a tiny sliver around the low-dimensional generator
  support (a region which has vanishingly small probability under the data distribution)
  without learning anything about the natural data distribution.
  In general the critic is incentivized to focus on tiny details relevant to detecting
  the current generator support but potentially imperceptible to a human, and the generator
  is incentivized to change these tiny details slightly to move the current support.
\end{itemize}
These pathologies seem highly undesirable for a probabilistic model.
Much attention has been devoted to this issue, especially under the assumption that the
data support is also low-dimensional
\cite[for example]{arjovsky_wasserstein_2017,mescheder2018training},
and it is one of the motivating scenarios for Wasserstein GANs.

This issue is extremely easy to fix by injecting noise at all levels of the generator
network, including the output (with a learned variance parameter).
If a given injection of noise is not useful then it easy enough for the generator to learn
to ignore it.
This injected noise (indeed, even just the output noise) is enough to formally make the
generator support cover all of output space and eliminate all of the above theoretical
pathologies.
The dimensionality argument above no longer applies since the dimensionality of all noise
sources is now greater than the output dimensionality (the output noise alone ensures this).
The injected noise can also have advantages in practice \citep{karras2019style}.

\clearpage
\renewcommand*{\bibname}{References}
\bibliographystyle{abbrvnat}
\bibliography{main}
\end{document}